\documentclass[11pt]{article}
\usepackage{acl2001}
\usepackage{epsf}
\title{Using a Support-Vector Machine
for Japanese-to-English Translation of Tense, Aspect, and Modality}
\author{Masaki Murata, Kiyotaka Uchimoto, Qing Ma, and Hitoshi Isahara \\
Communications Research Laboratory\\
2-2-2 Hikaridai, Seika-cho, Soraku-gun, Kyoto 619-0289, Japan\\
{\tt \{murata,uchimoto,qma,isahara\}@crl.go.jp}}

\begin{document}
\maketitle
\begin{abstract}
This paper describes experiments 
carried out using a variety of 
machine-learning methods, including the k-nearest neighborhood method 
that was used in a previous study, 
for the translation of tense, aspect, and modality. 
It was found that the support-vector machine method 
was the most precise of all the methods tested. 
\end{abstract}

\baselineskip=0.97\baselineskip

\section{Introduction}

Tense, aspect, and modality are known to cause problems 
in machine translation. 
In traditional approaches 
tense, aspect, and modality have been translated using 
manually constructed heuristic rules. 
Recently, however, corpus-based 
approaches such as the example-based method (k-nearest neighborhood method) have also been 
used \cite{Murata_modal_tmi}. 
For our study, 
we carried out experiments on 
the translation of tense, aspect, and modality 
by using a variety of machine-learning methods, 
in addition to the k-nearest neighborhood method, and 
then determined which method was the most precise. 
In our previous research, 
in which we studied the utilization of the k-nearest neighborhood method, 
only the strings at the ends of sentences were used 
to translate tense, aspect, and modality. 
In this study, 
however, we used all of the morphemes 
from each of the sentences 
as information, as well as used 
the strings at the ends of the sentences. 

In connection with our approach, 
we would like to emphasize the following points: 
{
\begin{itemize}
\item 
  We obtained better translation of 
  tense, aspect, and modality 
  by using a support-vector machine method 
  than we could by using the k-nearest neighborhood method 
  that we used in our previous study \cite{Murata_modal_tmi}.

\item 
  In our previous study \cite{Murata_modal_tmi} 
  we only used the strings 
  at the ends of sentences to translate 
  tense, aspect, and modality. 
  Here, however, 
  we used all of the morphemes in each of the sentences 
  as information, in addition to 
  the strings at the ends of the sentences. 
  Using a statistical test, 
  we were able to confirm that adding the morpheme information 
  from each of the sentences was 
  significantly effective in improving the precision of the translations. 
\end{itemize}}

\begin{figure}[t]
\small
  \begin{center}
    \leavevmode
\hspace*{-0.6cm}
\begin{tabular}{|p{7.7cm}|}\hline
g この子どもはああ言えばこう言うから小憎らしい.\\
This child always talks back to me, and this \verb+<v>+is\verb+</v>+ why I hate him.\\[0.2cm]
d 彼がああおくびょうだとは思わなかった.\\
I \verb+<v>+did not think\verb+</v>+ he was so timid.\\[0.2cm]
c ああ忙しくては休む暇もないはずだ.\\
Such a busy man as he \verb+<v>+cannot have\verb+</v>+ any spare time.\\\hline
\end{tabular}
\caption{Part of the modality corpus}
\label{fig:modal_corpus}
\end{center}
\end{figure}

\section{Task Descriptions}
\label{sec:mondai_settei}

For this study we used the modality corpus 
described in one of our previous papers \cite{NLP2001_eng}. 
Part of this modelity corpus is shown in Figure \ref{fig:modal_corpus}. 
It consists of a Japanese-English bilingual corpus, and 
the main verb phrase in each English sentence is tagged with \verb+<v>+. 
The symbols placed at the beginning of 
each Japanese sentence, such as ``c'' and ``d,'' indicate 
categories of tense, aspect, and modality 
for the sentence. 
(For example, ``c'' and ``d'' indicate
``can'' and past tense, respectively.)

\begin{table}[t]
\caption{Occurrence rates of categories}
\label{tab:distribution}
  \begin{center}
\small\renewcommand{\arraystretch}{1.000}
\begin{tabular}[c]{|l|c|c|}\hline
\multicolumn{1}{|c|}{Category} & Kodansha & White paper\\\hline
present    &  0.42  & 0.41 \\
past       &  0.36  & 0.21 \\
imperative &  0.05  & 0.00 \\
perfect    &  0.04  & 0.11 \\
``will''   &  0.03  & 0.06 \\
progressive&  0.03  & 0.10 \\
``can''    &  0.02  & 0.04 \\
others     &  0.05  & 0.07 \\\hline
\end{tabular}
\end{center}
\end{table}

The following categories were used for tense, aspect, and modality. 
{\small\begin{enumerate}
\item 
  All combinations of 
  each auxiliary verb (``be able to,'' ``be going to,'' ``can,'' ``have to,'' 
  ``had better,'' ``may,'' ``must,'' ``need,'' ``ought,'' ``shall,'' ``used to,'' and ``will'') 
  and forms for \{present tense, past tense\}, 
  \{progressive, non-progressive\}, \{perfect, non-perfect\}
  ($2^{15}$ categories)
\item 
  Imperative mood (1 category)
\end{enumerate}}
These categories of 
tense, aspect, and modality are defined on the basis of 
the surface expressions of the English sentences. 
Therefore, if we are able to determine the correct category from a Japanese sentence, 
we should also be able to translate the Japanese tense, aspect, and modality 
into English. 
In this study, 
only the tags indicating the categories of tense, aspect, and modality 
and the Japanese sentences were used. 

The following two types of corpora were used 
to construct the modality corpus. 

{\small\begin{itemize}
\item 
  Example sentences in the Kodansha Japanese-English dictionary 

  (39,660 sentences, 46 categories)

\item 
  White papers

  (5,805 sentences, 30 categories)

\end{itemize}}
The occurrence rates of major categories 
are shown in Table \ref{tab:distribution}. 
As can be seen in the table, the present tense occurs most frequently. 

\section{Machine-Learning Methods}
\label{sec:ml_method}

We used the following four machine-learning method for our study.\footnote{Although 
there are also decision-tree learning methods such as C4.5, 
we did not use them for the following two reasons. 
First, a decision-tree learning method performs worse than 
other methods for several tasks \cite{murata_coling2000,taira_svm_eng}. 
Second, the number of attributes used in this study 
was too large, and 
the performance of C4.5 would become even worse if 
the number of attributes was decreased 
so that it could be used.}
{\baselineskip=0.5\baselineskip
\begin{itemize}
\item 
  k-nearest neighborhood method
\item 
  decision-list method
\item 
  maximum-entropy method
\item 
  support-vector machine method
\end{itemize}}
In this next section, 
we will be explaining each of these machine-learning methods. 

\subsection{k-nearest neighborhood method}

The domain of machine translation includes
a method called an example-based method. 
In this method, 
the example most similar to the input sentence 
is searched for, and the category of 
the input sentence is chosen based on the example. 
However, this method only uses one example, so 
it is weak with respect to noise (i.e. errors in the corpus 
or other exceptional phenomenon). 
The k-nearest neighborhood method prevents this problem 
by using the most similar examples 
(a total of k examples) instead of 
using only the most similar example. 
The category is chosen 
on the basis of ``voting''\footnote{``Voting'' 
means a decision made by the majority.} on k examples. 
Since this method uses multiple examples, 
it is capable of providing a stable solution 
even if a corpus includes noise. 

In the k-nearest neighborhood method, 
since it is necessary to collect similar examples, 
it is also necessary to define the similarity 
between each pair of examples. 
The definition of similarity used in this paper 
is discussed in the section on features 
(Section \ref{sec:sosei}). 
When there is an example that has the same similarity 
as the selected k examples, 
that example is also used in the ``voting.'' 

\subsection{Decision-List Method}

In this method, 
the probability of each category is calculated 
using one feature $f_j (\in F, 1\leq j\leq k)$, and 
the category with the highest probability is judged to be 
the correct category. 
The probability that 
produces a category $a$ in a context $b$ is given 
by the following equation: 
{
\begin{eqnarray}
  \label{eq:decision_list}
  p(a|b) = p(a|f_{max}),
\end{eqnarray}
}
where $f_{max}$ is defined as
{
\begin{eqnarray}
  \label{eq:decision_list2}
  f_{max} = argmax_{f_j\in F} \ max_{a_i\in A} \ \tilde{p}(a_i|f_j),
\end{eqnarray}
}
such that $\tilde{p}(a_i|f_j)$ is the occurrence rate of 
category $a_i$ when the context includes a feature $f_j$. 

The decision-list method is simple. 
However, 
since it estimates the category on the basis of only one feature, 
it is a poor machine-learning method.  

\subsection{Maximum-Entropy Method}

In this method, 
the distribution of probabilities $p(a,b)$ 
when Equation (\ref{eq:constraint}) is satisfied and 
Equation (\ref{eq:entropy}) is maximized 
is calculated, and 
the category with the maximum probability 
as calculated from the 
distribution of probabilities is judged to be 
the correct category \cite{ristad97,ristad98}: 

{\footnotesize
\begin{eqnarray}
  \label{eq:constraint}
  \sum_{a\in A,b\in B}p(a,b)g_{j}(a,b) 
  \ = \sum_{a\in A,b\in B}\tilde{p}(a,b)g_{j}(a,b)\\
  \ for\ \forall f_{j}\ (1\leq j \leq k) \nonumber
\end{eqnarray}
}
{\footnotesize
\begin{eqnarray}
  \label{eq:entropy}
  H(p) & = & -\sum_{a\in A,b\in B}p(a,b)\ log\left(p(a,b)\right),
\end{eqnarray}}
where $A, B,$ and $F$ are a set of categories, a set of contexts, 
and a set of features $f_j (\in F, 1\leq j\leq k)$, respectively; 
$g_{j}(a,b)$ is a function with a value of 1 when context $b$ includes feature $f_j$ 
and the category is $a$, and 0 otherwise; 
and $\tilde{p}(a,b)$ is the occurrence rate of 
a pair $(a,b)$ in the training data. 

In general, the distribution of $\tilde{p}(a,b)$ is very sparse. 
We cannot use the distribution directly, 
so we must estimate the true distribution of $p(a,b)$ 
from the distribution of $\tilde{p}(a,b)$. 
With the maximum-entropy method, 
we assume that the estimated value of the frequency of 
each pair of categories and features 
calculated from $\tilde{p}(a,b)$ is the same as 
that calculated from $p(a,b)$, which corresponds to Equation \ref{eq:constraint}. 
These estimated values are not so sparse. 
We can thus use the above assumption to calculate $p(a,b)$. 
Furthermore, we maximize the entropy 
of the distribution of $p(a,b)$ to 
obtain one solution for $p(a,b)$, 
because using only Equation \ref{eq:constraint} produces 
several solutions for $p(a,b)$. 
Maximizing the entropy makes 
the distribution more uniform, which 
is known to provide a strong solution for data sparseness problems. 

\begin{figure}[t]
      \begin{center}
      \epsfile{file=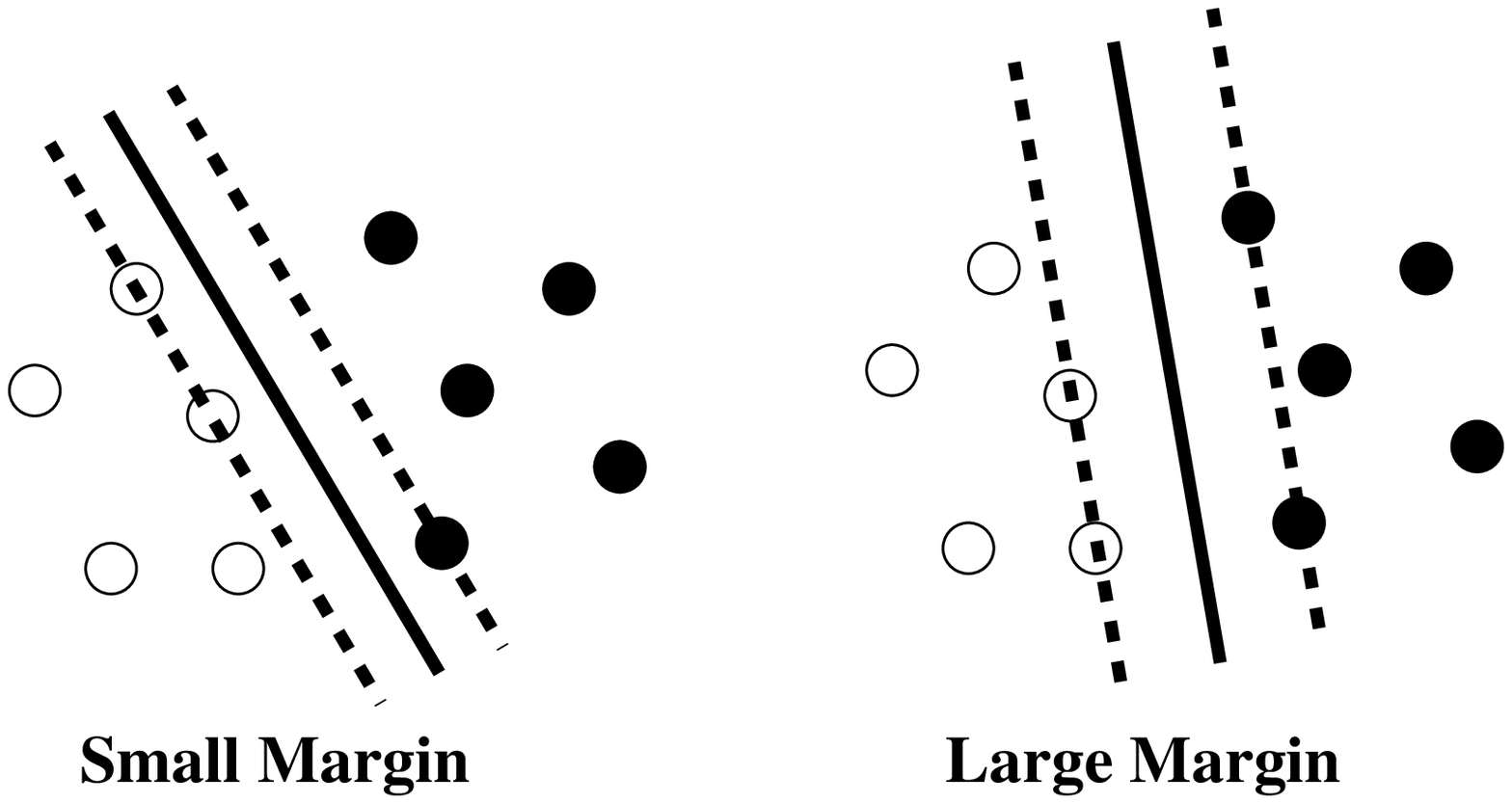,height=3cm,width=6cm} 
      \end{center}
      \caption{Maximizing the margin}
    \label{fig:margin}
\end{figure}

\subsection{Support-Vector Machine Method}

In this method, 
data consisting of two categories is classified 
by dividing space with a hyperplane. 
When the two categories are positive and negative and 
the margin between the positive 
and negative examples in the training data 
is larger (see Figure \ref{fig:margin}\footnote{In the figure, 
the white and black circles indicate 
positive and negative examples, respectively. 
The solid line indicates the hyperplane that divides the space 
and the broken lines indicate the planes at 
the boundaries of the margin regions.}), 
the possibility of incorrectly choosing 
categories in open data is thought to be smaller. 
The hyperplane that maximizes the margin is determined, 
and classification is carried out by uisng the hyperplane. 
Although the basics of this method are as described above, 
for the extended versions of the method 
the inner region of the margin in the training data 
can include a small number of examples, and 
the linearity of the hyperplane can be changed to non-linearity 
by using kernel functions. 
Classification in the extended method is equivalent to 
classification carried out using the following discernment function, and 
the two categories can be classified 
on the basis of whether the output value of the function is positive 
or negative \cite{SVM,kudoh_svm}: 

{\small
\begin{eqnarray}
  \label{eq:svm1}
  f({\bf x}) & = & sgn \left( \sum^{l}_{i=1} \alpha_i y_i K({\bf x}_i,{\bf x}) + b \right)\\
  b & = & -\frac{max_{i,y_i=-1}b_i + min_{i,y_i=1}b_i}{2}\nonumber\\
  b_i & = & \sum^l_{j=1} \alpha_j y_j K({\bf x}_j,{\bf x}_i), \nonumber
\end{eqnarray}
}
where ${\bf x}$ is the context (a set of features) 
of an input example, 
${\bf x}_{i}$ and $y_i \ (i=1,...,l, y_i\in\{1,-1\})$ indicate 
the context of the training data and its category, and the function $sgn$ is 
{\small
\begin{eqnarray}
  \label{eq:svm2}
  sgn(x) \, = & 1 & (x \geq 0),\\
          & -1 & (otherwise). \nonumber
\end{eqnarray}
}
Each $\alpha_i \ (i=1,2...)$ is fixed 
as the value of $\alpha_i$ 
when the value of $L(\alpha )$ in Equation (\ref{eq:svm4}) 
is at its maximum under the conditions of 
Equations (\ref{eq:svm5}) and (\ref{eq:svm6}). 
{\small
\begin{eqnarray}
  \label{eq:svm4}
  L({\alpha}) & = & \sum^l_{i=1} \alpha_i - \frac{1}{2} \sum^l_{i,j=1} \alpha_i \alpha_j y_i y_j K({\bf x_i},{\bf x_j})
\end{eqnarray}
}
{\small
\begin{eqnarray}
  \label{eq:svm5}
  0 \leq \alpha_i \leq C \, \, (i=1,...,l)
\end{eqnarray}
}
{\small
\begin{eqnarray}
  \label{eq:svm6}
  \sum^l_{i=1} \alpha_i y_i = 0 
\end{eqnarray}
}

Although the function $K$ is called a kernel function and 
various types of kernel functions are used, 
we used the following polynomial function: 

{\small
\begin{eqnarray}
  \label{eq:svm3}
  K({\bf x},{\bf y}) & = ({\bf x}\cdot{\bf y} + 1)^d.
\end{eqnarray}
}
$C$ and $d$ are constants set by experimentation, 
and in this paper, 
$C$ is fixed as 1 for all of the experiments. 
Two values, $d=1$ and $d=2$, are used for $d$. 
A set of ${\bf x}_i$ that satisfies $\alpha_i > 0$ is called 
a support vector, and 
the portion performing the sum in Equation (\ref{eq:svm1}) 
is calculated using only examples that are support vectors. 

Support-vector machine methods are capable of handling data 
consisting of two categories. 
In general, 
data consisting of more than two categories is handled 
using the pair-wise method \cite{kudoh_chunk_conll2000}. 

In this method, 
for data consisting of N categories, 
pairs of two different categories (N(N-1)/2 pairs) 
are constructed. 
The better category is determined 
using a 2-category classifier. In this paper, 
a support-vector machine\footnote{We used the software 
TinySVM \cite{kudoh_svm} by Kudo as a support-vector machine.}
was used as the 2-category classifier. 
Finally, the correct category is determined 
on the basis of ``voting'' on the N(N-1)/2 
pairs analyzed by the 2-category classifier. 

The support-vector machine method used in this paper was 
performed by combining the support-vector machine method 
and the pair-wise method described above. 

\section{Features (information used in classification)}
\label{sec:sosei}

Although we have already explained the four machine-learning methods 
in the previous section, we must also 
define the features 
(information used in classification). 
In this section, we will explain these features. 

As mentioned in Section \ref{sec:mondai_settei}, 
when a Japanese sentence is input, 
we then output the category of the tense, aspect, and modality. 
Therefore, features are extracted 
from the input Japanese sentence. 

\begin{table*}[t]
\caption{Precisions for the Kodansha data (values in parentheses are for the closed experiments)}
\label{tab:kodansha_result}
  \begin{center}
\small\renewcommand{\arraystretch}{1.000}
\begin{tabular}[c]{|l|cc|cc|cc|}\hline
\multicolumn{1}{|c|}{Method}  & \multicolumn{2}{|c|}{Feature-set 1} & \multicolumn{2}{|c|}{Feature-set 2} & \multicolumn{2}{|c|}{Feature-set 3} \\\hline
knn (k=1)         & ---     & ---       & 79.36\% & (98.50\%) & ---     & ---       \\
knn (k=3)         & ---     & ---       & 80.35\% & (83.94\%) & ---     & ---       \\
knn (k=5)         & ---     & ---       & 80.43\% & (82.39\%) & ---     & ---       \\
knn (k=7)         & ---     & ---       & 80.39\% & (81.71\%) & ---     & ---       \\
knn (k=9)         & ---     & ---       & 80.22\% & (81.30\%) & ---     & ---       \\
decision list     & 74.19\% & (98.21\%) & 80.23\% & (98.18\%) & 67.90\% & (86.58\%) \\
max. ent.         & 80.37\% & (88.87\%) & 81.16\% & (83.85\%) & 75.35\% & (84.15\%) \\
support vec. (d=1)& 82.48\% & (98.70\%) & 81.93\% & (98.50\%) & 78.68\% & (96.68\%) \\
support vec. (d=2)& 82.28\% & (98.48\%) & 81.37\% & (98.48\%) & 79.01\% & (98.74\%) \\\hline
\end{tabular}
\end{center}

\small\hspace*{2cm}
baseline = 73.88\%.
\end{table*}

We tested the following three kinds of feature sets 
in our experiments. 
{\small
\begin{itemize}
\item 
  {\bf Feature-set 1}

  Feature-set 1 consists of 1-gram to 10-gram strings at 
  the ends of the input Japanese sentences and 
  all of the morphemes from each of the sentences. 

  e.g. ``しない'' (do not), ``今日'' (today).
  
  (The number of features is 230,134 in the Kodansha Japanese-English dictionary 
  and 25,958 in the white papers.)

\item 
  {\bf Feature-set 2}

  Feature-set 2 consists of 1-gram to 10-gram strings at 
  the ends of the input Japanese sentences. 

  e.g. ``しない'' (do not), ``しなかった'' (did not). 

  (The number of features is 199,199 in the Kodansha Japanese-English dictionary 
  and 16,610 in the white papers.)

\item 
  {\bf Feature-set 3}

  Feature-set 3 consists of 
  all of the morphemes from each of the sentences. 

  e.g. ``今日'' (today), ``私'' (I), ``は'' (topic-marker particle) ``走る'' (run). 

  (The number of features is 30,935 in the Kodansha Japanese-English dictionary 
  and 9,348 in the white papers.)

\end{itemize}}
The Japanese morphological analyzer JUMAN \cite{JUMAN3.5e} 
was used to divide the input sentences into morphemes. 

Feature-set 1 is the combination of 
Feature-sets 2 and 3. 
Feature-set 2 was constructed based on our 
previous research \cite{Murata_modal_tmi}. 
In Japanese sentences 
the tense, aspect, and modality are 
often indicated by the verbs at the ends of sentences.\footnote{The Japanese 
language is of the type SOV, so verb phrases 
appear at the ends of sentences.}
Therefore, 
in our previous study, 
the strings at the ends of the sentences were used as features. 
Feature-set 3 was constructed by taking into consideration 
the fact that 
adverbs such as ``tomorrow'' and ``yesterday'' 
can also indicate tense, aspect, and modality, and 
must therefore be used. 

Defining the feature sets 
is sufficient for enabling the use of 
decision-list, 
maximum-entropy, and 
support-vector machine methods. 
For the k-nearest neighborhood method, 
however, 
it is also necessary to define the similarities between examples, 
in addition to the feature sets. 
For Feature-sets 1 and 3, which use 
all of the morphemes from the entire input sentence, it is 
difficult to define the similarity. 
Therefore, we decided to only use Feature-set 2 for 
the k-nearest neighborhood method. 
In terms of defining similarity for Feature-set 2, 
when two examples match for x-gram characters, 
the value of the similarity between them is x.

\section{Experiments}
\label{sec:experiment}

This section describes 
our experiments on the translation of tense, aspect, and modality 
that were conducted 
using the machine-learning methods described in Section \ref{sec:ml_method} 
with the feature sets described in Section \ref{sec:sosei} 
for the tasks described in Section \ref{sec:mondai_settei}. 

First, we conducted experiments using 
the example sentences in the Kodansha Japanese-English dictionary. 
The results for these experiments are shown in Table \ref{tab:kodansha_result}. 
We conducted two types of experiments, closed and open\footnote{
{\it Closed} means experiments 
that uses the tested data when learning. 
{\it Open} means experiments 
that do not use the tested data when learning.}. 
The open experiments were performed 
using 10-fold cross-validation.  
In the table, the values in parentheses 
indicate the precisions for the closed experiments 
and the values outside the parentheses are for 
the precisions for the open experiments.  
(We used the baseline method for comparison. 
This method is used to judge cases in which 
the end of the sentence is ``た,'' which 
is a Japanese particle used for the past tense, 
as the past tense, and judges 
other cases to be the present tense.)

We were able to learn the following from the experimental results. 
\begin{itemize}
\item 
  The cases of $k>1$ performed better than the case of $k=1$, 
  which is the example-based method. 
  We thus found that 
  the k-nearest neighborhood method was more precise than 
  the example-based method. 
  (This had also been confirmed in our previous research \cite{Murata_modal_tmi}.)

\item 
  The decision-list method had 
  almost the same precisions as the k-nearest neighborhood method 
  when Feature-set 2 was used. 

\item 
  The maximum-entropy method was more precise 
  than both the k-nearest neighborhood and 
  decision-list methods. 

\item 
  The support-vector machine method obtained 
  higher precisions than all the other methods. 

\item 
  In terms of comparing feature sets, 
  the maximum-entropy and decision-list methods 
  obtained their highest precisions when 
  Feature-set 2 was used. 
  They produced lower precisions 
  when morpheme information was added, as in Feature-set 1. 
  This would be because 
  the number of unnecessary features increases 
  as the total number of features increases. 

\item 
  In terms of comparing feature sets 
  for the support-vector machine method, 
  Feature-set 1 obtained the highest precisions. 
  This indicates that 
  adding the morpheme information was effective 
  in improving precision. 
  Since adding the morpheme information produced 
  lower precisions for the other methods, 
  we assume that 
  the support-vector machine method is more 
  capable of eliminating unnecessary features and 
  selecting effective features than 
  the other methods. 
  We can provide the following 
  explanations 
  for these results 
  based on the theoretical aspects of each of the methods. 

  {\small\begin{itemize}
  \item 
    Because the decision-list method chooses the category 
    by using only one feature, 
    it is likely that only one unnecessary feature will be used, 
    and that the precisions are likely to decrease 
    when there are many unnecessary features. 

  \item 
    Because the maximum-entropy method always uses 
    almost all of the features, 
    the precisions decrease 
    when there are many unnecessary features. 

  \item 
    However, 
    because the support-vector machine includes a function 
    that eliminates examples, 
    such that it only uses examples that are support vectors 
    and does not use any other examples, 
    it eliminates many unnecessary features along 
    with these examples. 
    The precisions are thus unlikely to decrease 
    even if there are many unnecessary features. 
    
  \end{itemize}}

\item 
  The method with the higest precision 
  among all the methods was the support-vector machine method 
  using $d=1$ and Feature-set 1. 

\end{itemize}

To confirm that 
using Feature-set 1 was better than 
using Feature-set 2, 
(in other words, 
to confirm that adding the morpheme information was effective) 
we conducted a sign test. 
This was done for the case of $d=1$, 
which produced better results 
than $d=2$. 
Among all of the 39,660 examples, 
the number for which the category was chosen 
incorrectly with Feature-set 1 
and correctly with Feature-set 2 was 648. 
For the opposite case (chosen 
incorrectly with Feature-set 2 
and correctly with Feature-set 1), the number was 427. 
We performed the sign test by using this statistical data 
and obtained results showing that 
a significant difference existed 
at a significance level of 1\%. 
We can thus be almost completely sure that 
adding the morpheme information was effective. 

\begin{table}[t]
\caption{Effective morpheme features}
\label{tab:keitaiso_sosei}
  \begin{center}
\small\renewcommand{\arraystretch}{1.000}
\begin{tabular}[c]{|r|l|}\hline
\multicolumn{1}{|c|}{Frequency}  & \multicolumn{1}{|c|}{Morpheme feature} \\\hline
  221   &  が (subject-case particle)    \\
  121   &  いる (is doing) \\
   89   &  ない (is not)\\
   28   &  ます (do) \\
   23   &  なら (if) \\
   23   &  きた (came)\\
   22   &  もう (already or yet)\\
   19   &  中   (between)\\
   18   &  最近 (recently)\\
   16   &  だろう (will)\\
   16   &  ました (did)\\
   14   &  まだ   (yet)\\
   13   &  れる   (can)\\
   12   &  なければ (if not)\\
   11   &  ましょう (let's)\\
   11   &  すっかり (entirely)\\
   10   &  られ (is done)\\
    8   &  あす (tomorrow)\\
    8   &  なんて (how)\\\hline
  \end{tabular}
\end{center}
\end{table}

\begin{table*}[t]
\caption{Precisions for the white papers (values in parentheses are for the closed experiments)}
\label{tab:hakusho_result}
  \begin{center}
\small\renewcommand{\arraystretch}{1.000}
\begin{tabular}[c]{|l|cc|cc|cc|}\hline
\multicolumn{1}{|c|}{Method}  & \multicolumn{2}{|c|}{Feature-set 1} & \multicolumn{2}{|c|}{Feature-set 2} & \multicolumn{2}{|c|}{Feature-set 3} \\\hline
Support Vec. (d=1)& 60.10\% & (99.81\%) & 56.61\% & (89.87\%) & 56.14\% & (96.67\%) \\
Support Vec. (d=2)& 64.67\% & (99.81\%) & 56.74\% & (89.87\%) & 62.07\% & (99.83\%) \\\hline
\end{tabular}
\end{center}

\small\hspace*{2cm}
baseline = 49.77\%.
\end{table*}

\begin{table*}[t]
\caption{Precisions for the Kodansha data and the white papers (values in parentheses are for the closed experiments)}
\label{tab:kodan_hakusho_result}
  \begin{center}
\small\renewcommand{\arraystretch}{1.000}
\begin{tabular}[c]{|l@{ }l|cc|cc|cc|}\hline
\multicolumn{2}{|c|}{Training data}  & \multicolumn{6}{|c|}{Test data}  \\\cline{3-8}
\multicolumn{2}{|c|}{}  & \multicolumn{2}{|c|}{Kodansha and White} & \multicolumn{2}{|c|}{Kodansha only} & \multicolumn{2}{|c|}{White papers only} \\\hline
Kodansha data&(d=1)         & 82.44\% & (98.71\%) & 82.48\% & (98.70\%) & 65.31\% & ---       \\
Kodansha data&(d=2)         & 82.31\% & (98.74\%) & 82.28\% & (98.48\%) & 51.92\% & ---       \\
White papers &(d=1)         & 60.02\% & (99.79\%) & 47.65\% & ---       & 60.10\% & (99.81\%) \\
White papers &(d=2)         & 64.01\% & (99.83\%) & 49.53\% & ---       & 64.67\% & (99.81\%) \\\hline
\end{tabular}
\end{center}
\end{table*}

Next, we examined which features were effective among 
all the features using the morpheme information. 
This was done by examining 
the features that appeared relatively frequently among the 648 examples 
for which the category was chosen incorrectly with Feature-set 1 
and correctly with Feature-set 2. 
We used a binomial test to choose an example whose occurrence rate 
among the 648 examples 
was significantly larger than among all the examples 
at a significant level of 1\% 
as a relatively frequently appearing example. 
The 20 most frequently occurring features are listed 
in Table \ref{tab:keitaiso_sosei}. 
As shown in the table, 
we were able to obtain features that are thought to be effective in 
determining tense, aspect, and modality, such 
as ``もう'' (already), ``最近'' (recently), ``だろう'' (will), ``まだ'' (yet), ``なければ'' (if not), ``ましょう'' (let's) and ``あす'' (tomorrow), 
and we believe that such features improved precisions. 

Next, we carried out experiments using the white papers. 
These experiments were performed using the support-vector machine method 
that produced good precisions for the Kodansha data. 
The open precisions were calculated using 
10-fold cross-validation 
in these experiments as well. 
The experimental results are shown in Table \ref{tab:hakusho_result}. 

We learned the following from the results. 
\begin{itemize}
\item 
  The highest precision for the white paper data was 
  64.67\%. 

\item 
  Feature-set 1 produced higher precisions than Feature-set 2. 
  Moreover, Feature-set 3 also produced higher precisions than Feature-set 2. 
  These results again confirmed that 
  adding the morpheme information for each of the sentences was effective 
  in improving precisions. 

\end{itemize}

We next performed experiments in which 
different-domain data was used as training data, 
such that the Kodansha data was used as the training data 
and the white paper data was used as the test data.\footnote{Sekine 
had carried out domain-dependent/ domain-independent experiments on parsing \cite{Sekine97_anlp}.}
We then examined how the precisions changed 
under these conditions. 
These experiments were performed using 
support-vector machine methods with d=1 or d=2, which 
both obtained good precisions. 
When the training data and the test data overlapped, 
10-fold cross-validation was used for the overlapping part. 
The experimental results are shown in Table \ref{tab:kodan_hakusho_result}. 

We learned the following from these results. 
\begin{itemize}
\item 
  When we used different-domain data as training data, 
  the precisions greatly decreased. 
  When the Kodansha data was analyzed 
  using the White paper data as training data or 
  the white paper data was analyzed 
  using the Kodansha data as training data, 
  the precisions decreased about 15\% 
  (82.48\% $\Rightarrow$ 65.31\% or 64.67\% $\Rightarrow$ 49.53\%). 

  We thus found that 
  using same-domain data is more effective 
  in terms of precision. 
  It is difficult to construct 
  a system adapted for different-domain data 
  with a method that uses hand-written rules. 
  However, 
  for methods using machine-learning, such as those 
  described in this paper, 
  since it is easy to change the training data to different-domain data and 
  then have the data learned again, 
  it is easy to construct a system adapted for different-domain data. 

\item 
  When both the Kodansha and the white paper data 
  were used as training data, 
  the precisions were almost the same or slightly decreased. 
  We thus found increasing 
  the size of training data is not always better and 
  adding different-domain data is not effective. 


\end{itemize}

\section{Conclusion}

Tense, aspect, and modality are known to present difficult problems 
in machine translation. 
In traditional approaches, 
tense, aspect, and modality have been translated using 
manually constructed heuristic rules. 
Recently, however, corpus-based 
approaches such as the example-based method (k-nearest neighborhood method) have also been 
applied \cite{Murata_modal_tmi}. 
We carried out experiments on 
the translation of tense, aspect, and modality 
by using a variety of machine-learning methods, 
as well as the k-nearest neighborhood method, and 
we determined which method was the most precise. 
In our previous research, in which we used the k-nearest neighborhood method, 
only the strings at the ends of sentences were used 
to translate tense, aspect, and modality. 
However, in this study 
we used all of the morphemes in each of the sentences 
as information, as well as 
the strings at the ends of each of the sentences. 

The support-vector machine method was found to produce the highest precisions 
of all the methods we tested. 
We were also able to obtained better translations of 
tense, aspect and modality 
than we could by using the k-nearest neighborhood method. 
Furthermore, we used a statistical test 
to confirm that 
adding the morpheme information for the entire sentence, which was 
not used in our previous study, was effective 
in improving precision. 
We also carried out experiments using 
a different-domain corpus. 
In these experiments, we confirmed that 
using a different-domain corpus as the training data 
produced very low precisions, and that 
we must construct a system for translating 
the tense, aspect, and modality for each domain. 
This also indicates that 
approaches using machine-learning methods, such as those described in this paper, 
are appropriate 
because it would be too difficult to construct systems 
adapted for different domains by hand.

{\small\baselineskip=0.9\baselineskip
\bibliographystyle{acl}
\bibliography{mysubmit}
}

\end{document}